\documentclass[journal,twoside,web]{ieeecolor}
\usepackage{generic}
\usepackage{cite}
\usepackage{amsmath,amssymb,amsfonts}
\usepackage{algorithmic}
\usepackage{graphicx}
\usepackage{algorithm,algorithmic}
\usepackage{multirow}
\usepackage{multicol}
\usepackage{enumitem}
\usepackage{paralist}
\usepackage{graphicx}
\usepackage{textcomp}
\newcommand{\ours}{General Healthcare Predictive Framework}
\newcommand{\oursarc}{GenHPF}
\usepackage{bm,amsmath}
\usepackage{xspace}
\usepackage{graphicx}
\usepackage{color}

\usepackage{colortbl}
\definecolor{light_gray}{RGB}{170,170,170}

\usepackage[utf8]{inputenc} 
\usepackage[T1]{fontenc}    
\usepackage{url}            
\usepackage{booktabs}       
\usepackage{amsfonts}       
\usepackage{nicefrac}       
\usepackage{microtype}      
\usepackage{longtable,lscape}
\usepackage{array,multirow}
\usepackage{enumitem}

\usepackage{wrapfig}

\newcommand{\hide}[1]{}

\newcommand{\mb}{\mathbf{m}}

\newcommand{\pb}{\mathbf{p}}

\newcommand{\vb}{\mathbf{v}}

\newcommand{\yb}{\mathbf{y}}

\begin{document}
\title{GenHPF: General Healthcare Predictive Framework for Multi-task Multi-source Learning}
\author{
Kyunghoon Hur,
Jungwoo Oh,
Junu Kim,
Jiyoun Kim,
Min Jae Lee,
Eunbyeol Cho,
Seong-Eun Moon,
Young-Hak Kim,
Louis Atallah,
Edward Choi
\thanks{Accepted at Journal of Biomedical and Health Informatics. Copyrights and Permissions Department, IEEE Publications Administrations
445 Hoes Lane, P.O. Box 1331, Piscataway, NJ 08855-1331.}
\thanks{Kyunghoon Hur,
Jungwoo Oh,
Junu Kim,
Jiyoun Kim,
Min Jae Lee,
Eunbyeol Cho, Edward Choi are with Kim Jaechul Graduate School of AI, Korea Advanced Institute of Science and Technology, Daejeon 34141, Republic of Korea (e-mail:pacesun@kaist.ac.kr; ojw0123@kaist.ac.kr; kjune0322@kaist.ac.kr; mjbooo@kaist.ac.kr;  eunbyeol.cho@kaist.ac.kr; jiyoun.kim@kaist.ac.kr; edwardchoi@kaist.ac.kr). 
}
\thanks{Seong-Eun Moon is with Naver AI Lab, Seongnam-si 13561, Republic of Korea (e-mail:seongeun.moon@navercorp.com). Young-Hak Kim is with Asan Medical Center, University of Ulsan College of Medicine, Republic of Korea (e-mail:mdyhkim@amc.seoul.kr). Louis Atallah is with Philips, Cambridge, Massachusetts, United States (e-mail:louis.atallah@philips.com). }
}

\maketitle

\begin{abstract}
Despite the remarkable progress in the development of predictive models for healthcare, applying these algorithms on a large scale has been challenging. 
Algorithms trained on a particular task, based on specific data formats available in a set of medical records, tend to not generalize well to other tasks or databases in which the data fields may differ.
To address this challenge, we propose \ours{} (\oursarc{}), which is applicable to any EHR with minimal preprocessing for multiple prediction tasks.
\oursarc{} resolves heterogeneity in medical codes and schemas by converting EHRs into a hierarchical textual representation while incorporating as many features as possible.
To evaluate the efficacy of \oursarc{}, we conduct multi-task learning experiments with single-source and multi-source settings, on three publicly available EHR datasets with different schemas for 12 clinically meaningful prediction tasks.
Our framework significantly outperforms baseline models that utilize domain knowledge in multi-source learning, improving average AUROC by 1.2\%P in pooled learning and  2.6\%P in transfer learning while also showing comparable results when trained on a single EHR dataset.
Furthermore, we demonstrate that self-supervised pretraining using multi-source datasets is effective when combined with \oursarc{}, resulting in a 0.6\%P AUROC improvement compared to models without pretraining.
By eliminating the need for preprocessing and feature engineering, we believe that this work offers a solid framework for multi-task and multi-source learning that can be leveraged to speed up the scaling and usage of predictive algorithms in healthcare.\footnote{Our code implementation is available on Github. \url{https://github.com/hoon9405/GenHPF}}
\end{abstract}


\section{Introduction}
\label{sec:introduction}
\IEEEPARstart{P}{atient} medical records which are regularly accumulated in the form of Electronic Health Records (EHR) have opened up new opportunities for data-driven models, which can improve the quality of patient care.
With the rapid adoption of artificial intelligence (AI) in healthcare, healthcare providers continue to develop models for different applications such as predicting patient outcomes~\cite{choi2016doctor, awad2017early, thiel2010early}, optimizing effective hospital operations~\cite{shameer2017predictive, ashfaq2019readmission} and diagnosing  diseases~\cite{landi2020deep, miotto2016deep, park2021development}.

Until now, traditional model development methods have been constrained by their reliance on task-specific feature engineering, wherein preprocessing techniques are predominantly tailored for individual tasks or applications. 
For instance, predictive modeling tasks for patient care or benchmarking, and quality improvement require this approach.
Consequently, each health system or research institute is compelled to employ its own data experts to meticulously preprocess medical records to suit specific tasks. This process can be time-consuming and expensive, ultimately restricting the range of potential applications~\cite{harutyunyan2019multitask}. 

Furthermore, this problem is exacerbated by the increasing number of tasks that require excessive overheads for the hospitals to develop and the managing of each task-specific model.
Moreover, the increasing number of tasks significantly burdens hospitals in terms of developing and managing task-specific models. For example, clinicians may need to simultaneously perform various prediction tasks, such as mortality and readmission, for the same patient.
To address this challenge, a comprehensive framework is required that can be applied to multiple tasks~\cite{alonso2016multitask} with minimal preprocessing, thereby minimizing the need for a meticulous design of input features.

This problem contributes to the inequality in healthcare AI, as algorithms are developed and used by large (typically academic data centers) with access to large data and research capabilities.
In reality, typical EHR datasets do not follow a single data format, particularly across geographies and multiple EMR providers. 
Each health system could store data according to its own needs, which consequently requires a level of manual harmonization. 

\begin{figure*}[h] 
    \begin{minipage}[c]{0.3\textwidth}
    \caption{
    The conventional approach for building predictive models uses domain-specific knowledge to preprocess data for each hospital (or health system) and task. 
    In contrast, our proposed framework uses text input features, eliminating the need for preprocessing and feature engineering specific to each hospital. 
    This allows us to train a unified model for two multi-source learning scenarios:
    1) Conventional supervised learning for multi-task learning and
    2) Self-supervised pretraining with unlabeled data.
    By employing transfer learning, our framework allows each trained model to conduct transfer learning in any hospital, irrespective of data format differences, thereby ensuring general adaptability across healthcare systems.
    }
    \label{fig:fig1}
  \end{minipage}\hfill
  \begin{minipage}[c]{0.65\textwidth}
    \includegraphics[width=\textwidth]{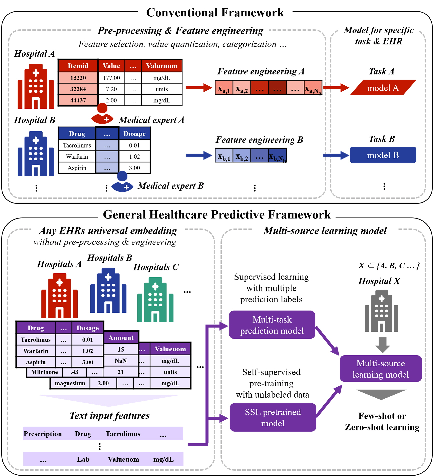}
  \end{minipage}
\end{figure*}

Specifically, different EHR systems adopt different medical code standards (\textit{e.g., ICD-9, ICD-10, raw text}), and use distinct database schemas to store patient records~\cite{johnson2016mimic, johnson2021mimic, pollard2018eicu}. 
\footnote{To unfold our tackling point conveniently, eICU is considered as EHR, which is originally collection across hospitals using different EHRs (EPIC, Cerner, etc.).}
These discrepancies in medical codes and schemas prevent healthcare institutions from conducting multi-source learning, such as fine-tuning a model that has been previously trained on a different EHR dataset (\textit{i.e., transfer learning}) or developing a unified model with data pooled from multiple hospitals (\textit{i.e., pooled learning}).

In summary, the major challenges encountered by current healthcare prediction models are as follows:
1)  models are specifically developed for each prediction task via feature engineering with task-specific domain knowledge, and 
2) procuring a large amount of unified data is difficult, which is a critical problem for developing the aforementioned general-purpose multi-task prediction model.
The main objective of this study is to propose a framework that addresses these two challenges.

\noindent{\textbf{Related work}} \label{sec:related work} \\
Previous healthcare prediction models with EHR have been focused on increasing the prediction performance by utilizing domain knowledge and various architectures such as recurrent neural networks (RNN)~\cite{lipton2015learning, choi2016doctor}, convolutional neural networks~\cite{nguyen2016mathtt}, and transformer-based models~\cite{vaswani2017attention, song2018attend, shang2019pretraining, choi2020learning}.
Although each study makes a distinct contribution, none address the two major aforementioned challenges.

\noindent{\textbf{Multi-Task Learning}}
MIMIC-Extract~\cite{wang2020mimicextract}, for example, performs domain-knowledge-based feature engineering, such as grouping semantically similar concepts into a clinical taxonomy as data structures that are directly usable in common multi-task time-series prediction pipelines.
Based on hand-crafted features, McDermott et al.~\cite{mcdermott2021a} proposed a benchmark for ten healthcare predictive tasks (multi-task learning) and reported their prediction performances. 
Because of their specialized nature, these approaches are designed to work exclusively for specific datasets, making them inapplicable to multiple EHR datasets that may vary in diversity and heterogeneity.

An alternative approach, proposed by Rajkomar et al.~\cite{rajkomar2018scalable}, involves a framework that incorporates all features of the EHR, that is, all column values in all the EHR tables. This allows the same model to be used for four different tasks. 
However, since this approach uses Fast Healthcare Interoperability Resources (FHIR)~\cite{mandel2016smart}, which is a form of Common Data Model (CDM), to manually standardize different EHR data into a uniform format, there is a significant overhead for multi-source learning. 
This process of standardizing EHR formats demands considerable domain knowledge and requires extensive manual efforts, making the integration of a large number of datasets into diverse formats impractical.

\begin{figure*}[ht]
\begin{minipage}[c]{0.3\textwidth}
    \caption{
    Overview of \oursarc. On the top, a patient's medical events occur over time.
    Each medical event $\mathcal{M}_i$ consists of event-related features $A_i^k$, including feature names and their values.
    These features, prepended with event type $e_i$, are converted to corresponding descriptions and tokenized into a sequence of sub-words.
    Then, an event encoder $f$ converts the sequence (\textit{i.e.}, event input) into an embedding $\mb_i$, which is then passed to the event aggregator $g$, which then makes a prediction $\hat{\yb}$
    .}
    \label{fig:fig2}
      \end{minipage}\hfill
      \begin{minipage}[c]{0.65\textwidth}
    {\includegraphics[width=\textwidth]{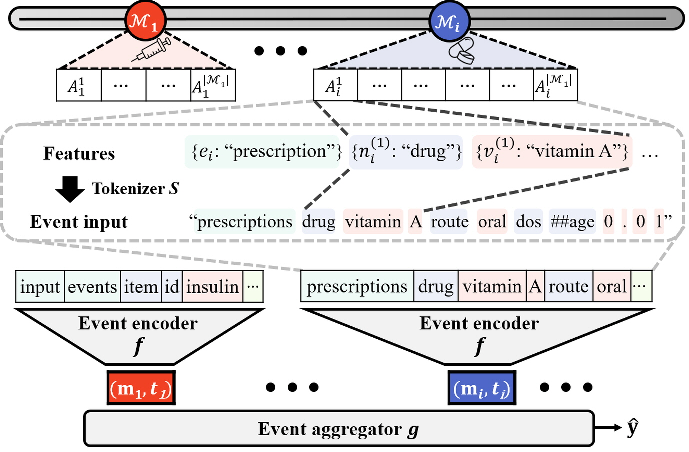}}
    \end{minipage}
\end{figure*}

\noindent{\textbf{Resolving EHR Heterogeneity without manual efforts}} 
To address the lack of scalability in previous works, AutoMap~\cite{wu2022automap} conducts medical code mapping via self-supervised learning using a predefined medical ontology.
This study aims to develop a solution to the current lack of a unified EHR system through a direct code-to-code mapping of two different medical institutions.
However, since AutoMap requires standardized medical ontology, manual efforts is still necessary. 

In another study, DescEmb~\cite{hur2022unifying} aimed to overcome the heterogeneity of medical codes by utilizing the clinical descriptions linked to each code, thereby partially enabling multi-source learning.
Despite its text-based embedding to avoid the manual code mapping process, this approach still necessitates domain experts to conduct EHR system-specific preprocessing to select compatible and meaningful features from the EHRs. 
Overcoming schema heterogeneity across different institutions poses a challenge when selecting universally applicable features with consistent formats from multiple datasets.
None of the aforementioned studies adequately address the dual challenges of utilizing multi-task models on heterogeneous EHRs.

\noindent{\textbf{Self-supervised pretraining in EHR}}
Self-supervised learning (SSL), which involves pretraining on large-scale unlabeled datasets and fine-tuning for prediction tasks, has demonstrated success in various applications~\cite{devlin2018bert, simclr, wav2vec} including predictive models based on EHR~\cite{rasmy2021medbert, zhang2022forecasting, chen2021disease, li2020behrt, steinberg2021language, yin2020tabert}. 
Previous studies on SSL using EHR data have primarily focused on pretraining and fine-tuning models exclusively for identical EHR systems, limiting their applicability to other EHR systems.
As the proposed framework resolves EHR heterogeneity, training it via SSL produces a general-purpose pretrained model that can be fine-tuned for any task in any EHR system.

This study makes three contributions.
To address both challenges (task-specific model development process and EHR heterogeneity) simultaneously, we propose \ours{} (\oursarc) (Figure~\ref{fig:fig1}), which is applicable to multiple patient record systems. 
GenHPF resolves heterogeneity in medical codes and schemas by converting medical records into a hierarchical textual representation while incorporating as many features as possible. 
This framework reflects the common data structure of medical records, allowing different structures to be utilized without code and schema harmonization processes.

Second, to demonstrate the efficacy of \oursarc{} empirically, we conduct extensive experiments using three publicly available EHR datasets with different schemas (MIMIC-III, eICU, MIMIC-IV) for the twelve clinically meaningful prediction tasks.
Our framework achieves comparable or higher prediction performances on single-domain learning compared with other frameworks, while consistently outperforming all other frameworks in terms of pooled learning and transfer learning.

Lastly, we combine several SSL methods with GenHPF, demonstrating the best practices that provides benefits to GenHPF as a self-supervised pretraining method with unlabeled data. 
This will enable researchers and engineers in this field to use a pretrained GenHPF as a general-purpose foundation model for diverse prediction tasks, regardless of the EHR schema.
Our findings provide insights for further research on the multi-source learning of EHR.
Figure~\ref{fig:fig1} overviews the proposed framework.

\section{Methodology} \label{sec:math}
\subsection{Structure of Electronic Health Records}
This section describes and summarizes the EHR structure and notations used throughout this paper.
In typical EHR data, each patient $P$ can be represented as a sequence of medical events $[\mathcal{M}_1,\ldots,\mathcal{M}_N]$, where $N$ is the total number of events throughout the entire patient visit history.
The i-th medical event of a patient $\mathcal{M}_i$ can be expressed as a set of event-associated features $\{A_i^{1},\ldots,A_i^{|\mathcal{M}_i|}\}$.
Each feature $A_i^{k}$ can be seen as a tuple of a feature name and its value $(n_i^{k}, v_i^{k}), n_i^k \in \mathcal{N}, v_i^k \in \mathcal{V}$, where $\mathcal{N}$ and $\mathcal{V}$ are each
a set of unique feature names 
(\textit{e.g.}, $\{$``drug name'', ``drug dosage'', $\ldots, \}$)
and feature values
(\textit{e.g.}, $\{$``vancomycin'', ``10.0'', $\ldots, \}$), respectively.

In addition, each medical event $\mathcal{M}_i$ has its corresponding event type $e_i \in \mathcal{E}$ which denotes the type of the event
(\textit{e.g.}, $\mathcal{E} = \{$``\textit{lab test}'', ``\textit{prescription}'', $\ldots, \}$).
Lastly, since the recorded time is also provided with $\mathcal{M}_i$, we can measure the time interval $t_i$ between $\mathcal{M}_i$ and $\mathcal{M}_{i+1}$.

\subsection{General Healthcare Predictive Framework}
In this section, we present \oursarc, a general framework for EHR-based prediction based on the following three principles, and describe how to implement each principle: 
(1) text-based embedding, (2) employing the entire features of EHR, and (3) medical event aggregation.
Figure~\ref{fig:fig2} depicts the overall architecture.

\noindent{\textbf{Text-based embedding.}}
A conventional EHR embedding method begins by assigning a unique embedding for each element in $\mathcal{V}$ via a linear map (\textit{i.e.}, lookup table) $f_{\mathcal{V}}$~\cite{choi2016multi, song2018attend, song2019medical, mcdermott2021a, rajkomar2018scalable}, so that $v_i^k$ can be converted to a vector $\vb_i^k \in \mathbb{R}^{d_v}$,
typically followed by pooling multiple feature values ($\vb_i^1, \vb_i^2, \ldots$) to obtain $\mb_i \in \mathbb{R}^{d_m}$, the embedding of $\mathcal{M}_i$.
\footnote{Previous EHR embedding methods do not typically use the feature name $n_i^k$}.
This conventional embedding, however, usually requires a different $f_{\mathcal{V}}$ for each medical institution due to the $\mathcal{V}$ \textit{heterogeneity}, namely each institution using different $\mathcal{V}$'s.
For example, MIMIC-III~\cite{johnson2016mimic}, an open-source EHR data, uses the ICD-9 diagnosis codes for recording diagnostic information, while eICU~\cite{pollard2018eicu}, another open-source EHR data, uses in-house diagnosis codes.
Therefore, the conventional embedding is not the most suitable foundation on which to build a general EHR framework.

DescEmb~\cite{hur2022unifying} proposed to resolve this problem by suggesting a text-based embedding, where hospital-specific feature values are first converted to textual descriptions (\textit{e.g.}, ``401.9'' $\rightarrow$ ``unspecified essential hypertension''),
then a text encoder paired with a sub-word tokenizer is used to obtain $\mb_i$~\cite{bojanowski2017enriching}.
 With this approach, the model can learn the language of the underlying medical text rather than memorize a unique embedding for each hospital-specific feature value, thereby overcoming the $\mathcal{V}$ \textit{heterogeneity} as the same text encoder can be used for all institutions that use the same language.
 At this point, we adopted this code-agnostic embedding method and extended it by utilizing feature names as well as the feature values, which is $s(t(n) + t(v))$ as the event representation.
 
We extend the previous approach by applying the text-based embedding philosophy to event types $e_i$ and feature names $n_i^k$, in addition to feature values $v_i^k$, as follows:
\begin{equation*}
    \mb_i = f \Big( S(e_i), S(n_i^1), S(v_i^1), \ldots, S(n_i^{|\mathcal{M}_i|}), S(v_i^{|\mathcal{M}_i|})
    \Big) \label{eq:event_embedding}
\end{equation*}
where $S$ is a sub-word tokenizer, and $f$ is an event encoder that takes a sequence of sub-word tokens and returns $\mb_i$.
Note that $f$ can be a pretrained language model as in DescEmb, or a randomly initialized transformer encoder, or even a single-layer RNN.
Although $f$ can be implemented with any sequence encoder (\textit{e.g.}, a pretrained language model as in DescEmb), we use 2-layer transformer in this work.


\noindent{\textbf{Employing the entire features of EHR.}\quad}
To develop a general predictive framework, in addition to the $\mathcal{V}$ \textit{heterogeneity}, we must consider the \textit{schema heterogeneity}, namely each medical institution using a different database schema. 
When developing a conventional predictive model, medical domain experts are typically involved to define $\mathcal{M}'_i \subset \mathcal{M}_i$, a subset of task-specific features among $\mathcal{M}_i$ according to each EHR system.
This process must be carried out repeatedly whenever they encounter a different EHR schema.
Moreover, in multi-source learning, medical domain experts must select and match compatible features between distinct EHR systems.
For instance, in the \textit{Lab} event of eICU, the feature named ``labResult'' should be paired with the ``VALUENUM'' feature in MIMIC-III's \textit{LABEVENTS} event.
Assessing database schemas of multiple sources and matching compatible features,
although inevitable in a conventional approach, is time-consuming and prone to human errors.

Therefore, to leverage multiple heterogeneous EHR sources, features that share the same meaning must be matched.
To avoid this costly procedure, our framework exploits the entire features of medical events, effectively resolving the schema heterogeneity.
As described in Eq.~\ref{eq:event_embedding}, the entire set of features in medical events is embedded into one unified embedding $\mb_i$. 
Since this approach utilizes all features, feature selection is not required.
Additionally, in multi-source learning, our framework is not constrained by the features that are present in each schema since both the name $n_i^k$ and the value $v_i^k$ of the feature are used.
A formal comparison of the conventional approach, DescEmb~\cite{hur2022unifying} and our approach for obtaining $\mb_i$ is provided below:
\begin{flalign}
& \mbox{\textit{Conventional approach}:} \nonumber && \\
& \quad \mb_i = pool (\{ f_{\mathcal{V}} (v_i^k) \mid A_i^k \in \mathcal{M}'_i\} ) \nonumber && \\
& \mbox{\textit{DescEmb}:}  && \nonumber \\
& \quad \mb_i = f \Big( \{ S(v_i^k) \mid A_i^k \in \mathcal{M}'_i \} \Big) \nonumber && \\
& \mbox{\textit{\oursarc}:}  && \nonumber \\
& \quad \mb_i = f \Big( S(e_i), \{ S(n_i^k), S(v_i^k) \mid A_i^k \in \mathcal{M}_i \} \Big) \nonumber
\end{flalign}

where \textit{pool} is typically implemented as a concatenation or summation of the elements.
Note that \oursarc{} differs from previous approaches in that it is the only approach to exploit all available information in a medical event, including the event type, all event names, and all event values.
Therefore, \oursarc{} provides a general solution applicable to any EHR system with a different schema, making it schema-agnostic, without requiring medical domain knowledge.
DescEmb~\cite{hur2022unifying} still cannot resolve this since it exploits only the feature value $v_i^{k}$. 
This approach does not take into account the need for the model to learn the semantics of column names, thereby necessitating only the selection of compatible features.

\noindent{\textbf{Medical event aggregation.}}
To leverage the EHR structure characteristics, where $P$ consists of a sequence of $\mathcal{M}_i$ and each $\mathcal{M}_i$ consists of a set of $A_i^k$, we design a hierarchical model consisting of the event encoder $f$, and the event aggregator $g$.

As each $\mathcal{M}_i$ is converted into $\mb_i$ according to Eq.~\ref{eq:event_embedding}, we can obtain $\pb \in \mathbb{R}^{d_p}$, the vector representation of $P$ as follows:
\begin{equation*}
    \pb = g \Big( (\mb_1, t_1),  (\mb_2, t_2),  \ldots, (\mb_N, t_N) \Big)  \label{eq:hierarchical}
\end{equation*}
where $g$ is an embedding function that takes a sequence of event embeddings, and t is a timestamp which is applied as following \cite{press2021train}, imposing the weight for attention according to the time interval between adjacent events.
Note that $g$ can be implemented with any sequence encoder, such as a Transformer encoder or a single-layer RNN.
Then, feeding $\pb$ through a softmax layer (sigmoid layer if binary prediction) will give us the final prediction $\hat{\yb}$.

In addition, $\pb$ can be obtained by employing a flattened model architecture rather than a hierarchical one, where sub-word tokens from all features of all medical events are passed to the sequence model $h$ at the same time.
We confirm that the hierarchical approach, which reflects the structure of EHR data, indeed outperforms the flattened approach.



\subsection{Self-supervised pretraining}
Building upon the premise that self-supervised pretraining may enhance downstream task performance, the proposed framework enbales multiple heterogeneous EHRs to be used during the self-supervised pretraining process. 
Our investigation focuses on determining the efficacy of various self-supervised pretraining approaches when applied to \oursarc{}.
In this study, we test four well-known SSL methods as follows:

\noindent{\textbf{SimCLR}~\cite{simclr}}: We execute a two-step process of (1) EHR data augmentation and (2) contrastive pretraining inspired by SimCLR.
For data augmentation, we create a pair of views per patient by halving the time-series data based on the number of events and randomly masking the tokens in the events at a fixed ratio.
The contrastive pretraining objective is to maximize the similarity of the representation vectors created from two views of the same patient \textit{(i.e., positive pair)} while minimizing the similarity of the vectors created from the views of different patients \textit{(i.e., negative pair)} in accordance with the SimCLR settings~\cite{simclr}.

\noindent{\textbf{Wav2Vec 2.0}~\cite{wav2vec}}: We execute the Wav2Vec 2.0~\cite{wav2vec} pretraining process, which consists of (1) feature encoder output quantization and (2) contrastive learning on mask-selected patient event timesteps.
During the quantization stage, continuous latent vectors (i.e., event encoder outputs) are quantized via mapping the vectors to discrete entries of a trainable codebook. 
Gumbel softmax is used to map each latent vector to the codebook entries. 
During the second stage, a proportion of the latent vectors are randomly masked before being fed into the event aggregator.
For each mask selected position, the overall pretraining objective is to maximize the similarity between the event representation vectors (i.e., event aggregator outputs) and their corresponding quantized vector, while minimizing the similarity with other quantized vectors.
The loss terms are followed as defined in Wave2Vec 2.0.
We use the event encoder as the feature encoder instead of the convolutional blocks used in the original study.

\noindent{\textbf{MLM and SpanMLM}~\cite{devlin2018bert, joshi2020spanbert}}:
For MLM pretraining, we randomly mask a fixed ratio of tokens among the whole patient event history, and the pretraining objective is to predict the masked tokens based on bidirectional attention.
For SpanMLM pretraining, we apply event-level random masking, where all tokens included in the sampled events are masked, which is intended to learn the context of the EHR time-series event by learning the event itself rather than simply learning the partial random masked sub-word of the description.
Note that both MLM and SpanMLM are based on predicting the raw text (\textit{i.e.} tokens), which prevents us from using the hierarchical textual representation (Figure 4). 
Therefore we use a flattened textual representation for these two methods; the Methods section describes this representation further.

\section{Experimental Settings and Design} \label{sec:results}
\subsection{Datasets}
\begin{table*}[ht]
    \caption{\label{tab:tabstat} Characteristics of datasets}
    \centering
    \begin{tabular}{ccccccc}
    \toprule
    Statitics                                                      & \multicolumn{2}{c}{MIMIC-III} & \multicolumn{2}{c}{eICU}  & \multicolumn{2}{c}{MIMIC-IV} \\ \hline \midrule
    No. of Observations                    & \multicolumn{2}{c}{38040}     & \multicolumn{2}{c}{98904} & \multicolumn{2}{c}{65511}    \\
    No. of ICU stay                        & \multicolumn{2}{c}{38040}     & \multicolumn{2}{c}{98904} & \multicolumn{2}{c}{65511}    \\
    Mean No. of events per sample          & \multicolumn{2}{c}{102.5}     & \multicolumn{2}{c}{48.5}  & \multicolumn{2}{c}{88.7}     \\ \hline
    Feature selection                                              & -              & O            & -            & O          & -             & O            \\ 
    \hline
    \midrule
    No. of Unique code                     & 10434          & 6370         & 6302         & 5704       & 9565          & 5808         \\
    No. of Unique subwords text            & 3321           & 2793         & 2678         & 2451       & 3512          & 3112         \\
    Mean No. features per event            & 7.2            & 4.5          & 6.7          & 5.2        & 10            & 4.4          \\
    Mean length of subwords text per event & 44.6           & 25.8         & 51           & 34.6       & 62.2          & 24.7         \\
    \bottomrule
    \end{tabular}
    \vskip -10pt 
\end{table*}
We use three publicly available datasets; MIMIC-III~\cite{johnson2016mimic}, MIMIC-IV~\cite{johnson2021mimic}, and eICU~\cite{pollard2018eicu}.
The MIMIC-III database consists of clinical data of over 40,000 patients admitted to the intensive care units (ICU) at the Beth Israel Deaconess Medical Center.
MIMIC-IV is an enhanced version of MIMIC-III that incorporates additional data sources, including admission date.
The eICU consists of ICU records from multiple US-based hospitals, with 140,000 unique patients.
All three datasets contain patient medical events including lab tests, prescriptions, and input events (\textit{e.g.}, drug injection), which are processed as inputs for the experiments.
Each event is marked with a timestamp.
We build patient cohorts of patients over the age of 18 years who remained in an ICU for over 24 hours.
To ensure reliable experiments and analyses, we randomly split each dataset into training, validation, and test sets in an 8:1:1 ratio.

Minimal preprocessing applicable to any EHR is performed in three steps. 
First, we eliminate features whose values consisted only of integers. 
This approach ensures that all continuous-valued features (e.g., lab test results) and textual features (e.g., lab test names) are used, while omitting features such as the patient ID. 
Second, we split numeric values digit by digit and assign a unique token to each digit place, a method known as \textit{digit place embedding} which was first introduced in DescEmb~\cite{hur2022unifying}.
Subsequently, we tokenize all features and prepare them as text input features using bio-clinical-bert tokenizer~\cite{alsentzer-etal-2019-publicly}.
Table~\ref{tab:tabstat} summarizes the general characteristics of the three datasets including the size and feature dimensions.
The embedding method for each feature is either a categorized feature (code-based embedding) or is the text itself.

For the pretraining dataset, we prepare an unlabeled dataset, employing multiple ICUs without an observation window, and sampled medical events with a maximum length of 150, except for the test set of the downstream task.
For medical sequences exceeding 150 events, we shift the starting point of sampling by 30 events, thereby altering the sample while maximizing data inclusion.

\subsection{Prediction tasks}
To fairly evaluate our framework for varous healthcare predictive tasks, we utilize open-source prediction tasks that can be applied in an ICU setting. 
We adopt eight prediction targets (Mort, LMort, Readm, Los3, Los7, Dx, Fi\_ac, and Im\_disch), as described by McDermott et al.~\cite{mcdermott2021a}. 
Additionally, to demonstrate the efficacy of GenHPF in a broader range of tasks, we formulate four prediction targets for lab values, which serve as proxy indicators for sepsis or acute kidney injuries~\cite{gupta2020sequential}.
All tasks are based on ICU stays, and the performance is evaluated using the area under the area under the receiver operating characteristic curve (AUROC).
Each task is defined as follows:
\begin{compactitem}
\item  \textit{Mortality} (Mort) (binary):
A sample is labeled positive for mortality if the discharge state was ``expired'' within a prediction window of 48 hours during the stay. 
In addition, for a longer-term prediction mortality prediction, we use death within 2 weeks (abbreviated as LMort).
\item  \textit{Length-of-Stay} (LOS) (binary):
The length of stay prediction for ICU stays can be categorized into two cases: determining whether a given stay lasted longer than 3 days (LOS3), and determining whether it lasted longer than 7 days (LOS7).
 \item \textit{Readmission} (Readm) (binary):
Given a single ICU stay, we consider a positive case of an ICU stays followed by another (readmission) during the same hospital stay.
\item  \textit{Final Acuity} (Fi\_ac) (multi-class):
Predicts the patient's discharge location at the end of their hospital stay, including patient expiration.
\item  \textit{Imminent Discharge} (Im\_disch) (multi-class):
Predicts whether the patient will be discharged within a prediction window of 48 h and if discharged, predicting the discharge destination. 
\item \textit{Diagnosis} (Dx) (multi-label):
Predicts all diagnosis (Dx) codes accumulated during an entire hospital stay.
We group Dx codes into 18 Dx classes using Clinical Classification Software (CCS) for the ICD-9-CM criteria [healthcare2016hcup]. 
\item \textit{Lab values} (multi-class):
Four distinct laboratory values[Creatinine (Crt), Bilirubin (Blr), Platelets (Plt))] are categorized into 5 classes, based on their corresponding ranges. These classes are derived using the thresholds employed to determine the Sequential Organ Failure Assessment (SOFA) scores. 
White blood cell (Wbc) is categorized into 3 classes.
\end{compactitem}

Using medical event information from the initial 12 hours after ICU admission, we apply a 12-hour timegap across all tasks. 
timegap, which is designed to exclude any data close to the prediction time, is implemented to maintain the task challenges and prevent potential data leakage. 
Excluding any ICU stays shorter than 24 h allows for both a 12-h observation window and a 12-h gap.
For diagnosis, we categorize diagnosis labels into 18 distinct classes based on the CCS ontology~\cite{healthcare2016hcup}. 
We use ICD9 for MIMIC-III, ICD10 for MIMIC-IV, and text format diagnostic labels for the eICU. 
We perform an additional mapping process for Fi\_ac and Im\_disch owing to the different labeling sources across datasets. 
For the lab value prediction tasks, we adopt the approach of Gyawali et al.~\cite{gyawali2019sepsis}, defining them from the SOFA score, which guides the severity of sepsis from specific lab values.
Hence, each lab value is assigned as a categorical value based on its corresponding SOFA score.
Statistics for prediction tasks are shown in Tables~\ref{tab:binary}, ~\ref{tab:multi-label}, ~\ref{tab:multi-class}.

\begin{figure*}[ht] 
    \centering
    \vskip -5pt
    \caption{Comparison of single domain learning and pooled learning prediction performances. 
    (A) Results of the average AUROC on 12 prediction tasks. 
    The data sources used for the evaluation are at the top of each graph. 
    The y-axis indicates the AUROC. 
    Each dot represents models (color) with source datasets used for training (shape) following the legends. 
    Note that ``Single'' refers to the same data source as the evaluation dataset. 
    The blue dashed line separates models into conventional embedding models (left- SAnD, Rajkomar) and text-based embedding models (right- DescEmb, \oursarc{}).
    Stars indicate the p-value of the t-test conducted to assess the significance between single-domain prediction and pooled learning. 
    (B) Results of each prediction task using MIMIC-III as the source dataset.
    }
    {\includegraphics[width=.85\linewidth]{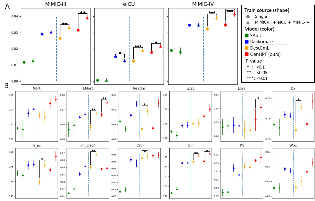}}
    \vskip -15pt
    \label{fig:fig3}
\end{figure*}

\subsection{Baselines and implementation details}
\noindent{\textbf{Baselines}}
As there is no previous work, to our knowledge, that tackled exactly the same goal as ours, we modified well-known general-purpose EHR embedding frameworks.
By comparing \oursarc{} with baselines, we systematically evaluate the components that can influence prediction performance in multi-source learning settings.
We analyzed these frameworks based on two options: feature utilization (selective or utilizing all) and embedding method (code-based or text-based).
In addition, all models were provided with both $n_i^k$ and $v_i^k$ for a fair comparison with \oursarc.

\begin{compactitem}
\item SAnD: This uses the conventional embedding, selected features $\mathcal{M}'_i$, and the flattened architecture, similar in spirit to SAnD~\cite{song2018attend}.
Note that feature embeddings from all medical events $[\mathcal{M}_1, \ldots, \mathcal{M}_N]$ are directly fed to the sequence encoder $h$ instead of being pooled to obtain individual $\mb_i$.
\item Rajkomar: This uses the conventional embedding, entire features $\mathcal{M}_i$, and the hierarchical approach, similar in spirit to \cite{rajkomar2018scalable} except the CDM standardization.
Note that the feature embeddings from each $\mathcal{M}_i$ are fed to $f$ to obtain individual $\mb_i$, which is then fed to $g$.
\item DescEmb: This uses the text-based embedding, selected features $\mathcal{M}'_i$, and the hierarchical approach, similar in spirit to DescEmb~\cite{hur2022unifying}.
\item AutoMap: This uses the same embedding method and features as Rajikomar~\cite{rajkomar2018scalable}. It trains $\mathcal{M}_i$ by automatically mapping medical codes using ontology-level alignment with an unsupervised learning method.
\item Muse: This uses the same embedding method and features as Rajikomar~\cite{rajkomar2018scalable}. It trains $\mathcal{M}_i$ using skip-grams and aligns the embedding space between bilingual dictionaries.

\end{compactitem}

\noindent{\textbf{Model implementation}}
For a fair comparison, $f$ and $g$ were both implemented with a randomly initialized 2-layer Transformer encoder, and $ç$ a 4-layer Transformer encoder, making all models equivalent in terms of the number of trainable parameters ($d_v = 128$, $d_m = 128$, $d_p = 128$).
\footnote{Note that we modified each baseline from the original frameworks for a fair comparison (e.g. transformer architecture~\cite{vaswani2017attention} which is a state-of-the-art model, is used instead of RNN.)}
Although all frameworks share the same sequence of medical events, the selection of features and the embedding approach employed can vary across each frameworks.
The selected features $\mathcal{M}'_i$ \footnote{For example, from the prescription event, we chose essential features such as drug name, drug volume, and unit of measurement among all available features.} followed by DescEmb~\cite{hur2022unifying}.

To maintain the same input information for both hierarchical and flattened models, we limit the number of events per sample.
Owing to computational resource constraints, the flattened models are limited to a maximum sequence length of 8192, and a correspondingly adjusted number of events were used as input for the hierarchical model, which includes the same events as the flattened model.

\noindent{\textbf{Training details}}
All experiments are conducted using five random seeds which are used to initialize the model parameters and to split the dataset.
Their performance is evaluated based on the area under the receiver operating characteristics (AUROC) averaged over twelve tasks.
We conduct all experiments in a multi-task learning setting, as our main interest is to develop a single model that performs multiple tasks using multiple EHR datasets simultaneously.
For multi-source learning, we train the combined dataset and validate each individual dataset separately.
Early stopping is enforced according to the validation AUROC for each dataset, and the best model is saved per dataset. 
Subsequently, each saved model is used to test the corresponding dataset.

\noindent{\textbf{Hyperparameters}}
We explored various hyperparameters to determine the optimal for each framework.
However, we found that the impact of these hyperparameters on the results was not significant.
Consequently, we use a unified set of hyperparameters for all cases, thereby simplifying the experiment while maintaining the performance for each model.
The final hyperparameters are a dropout of 0.3, a batch size of 64, and a learning rate of 1e-4.
For pretraining, we apply token masking with the same fixed ratio to SimCLR, MLM, and SpanMLM, in which 80\% of the randomly chosen token positions are replaced with the [MASK] token, 10\% of the positions are replaced with a random token, and the remaining 10\% of the positions are unmodified.
We apply Wav2Vec settings with 2 codebooks, 320 entries per codebook, a masking ratio of 65\%, and a feature gradient multiplication of 0.1 which slows down the event encoder gradient update.
For the codebook diversity loss weight, we use 0.1, 0.3, 0.1, 0.5 for MIMIC-III, eICU, MIMIC-IV, and the pooled domain, respectively.

\begin{figure*}[ht] 
 \centering
     \vskip -15pt
     \caption{Transfer learning results. 
    The source data used for training and target data used for evaluation with zero-shot or few-shot learning are indicated at the top of each graph.
    The source dataset is on the left side of the arrow, and the target is on the right.
    The y-axis indicates the AUROC and the x-axis is the portion of the target dataset for zero-shot or few-shot learning.
    Shading around the lines indicates the standard error from five seed experiments.
    For comparison with single domain performances, single domain learning performances of \oursarc{} are marked with the dashed line.} 
    {
    \includegraphics[width=0.8\linewidth]{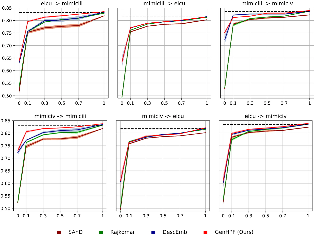}
    }
    \vskip -15pt
    \label{fig:fig4}
\end{figure*}

\subsection{Experimental Design}
To assess the efficacy of \oursarc{} in various aspects, we developed a series of prediction tasks across four distinct scenarios: (1) single-domain learning, (2) pooled learning, (3) transfer learning, and (4) self-supervised learning. For pooled learning and transfer learning, we follow the settings from DescEmb~\cite{hur2022unifying}.
For single-domain learning, models are trained and tested on a single dataset. This part tests \oursarc{} for single-domain learning although its primary aim is that of multi-source learning.
In pooled learning, it is crucial to utilize data collected from multiple EHR systems by leveraging the wealth of EHR data for prediction tasks.
Each framework simultaneously is trained on all three datasets, and evaluated separately on each dataset.
We compare the performance of single-domain learning and pooled learning to show that training on multiple datasets enhances predictive performance compared with models trained on a single dataset.

Next, in transfer learning, we aim to show that \oursarc{} can be beneficial when trained on a specific dataset and directly tested on other datasets (zero-shot learning) or when further trained on limited data (few-shot learning).
In practice, a single deep-learning model is typically trained on a large-scale hospital dataset and subsequently transferred to individual institutions, which could enable small hospitals to benefit from models trained on a large scale. 
Apart from acquiring large and representative datasets, this also entails ensuring compatibility between code and data schemas across different EHR systems, akin to what is necessary in pooled learning. 
In this scenario, each model is first trained on a source dataset and then directly evaluated on a sample from the same dataset(\textit{i.e.,} zero-shot) or further trained (\textit{i.e.,} fine-tune) on a target dataset. 

Finally, we investigate which SSL method with unlabeled data exhibits a performance improvement when fine-tuning the pretrained model on the prediction task.
To demonstrate the benefit of our approach, we compare three models: 1) a randomly initialized model trained on a single dataset; 2) a pretrained model and fine-tuned on a single dataset; 3) a pretrained model on the multi-source (pooled) dataset and fine-tuned on a single dataset.
Additionally, we assess the impact of pretraining on different fine-tuned data size settings, namely sample data, and full data, assuming that a pretrained model can be fine-tuned on a smaller hospital or a similar-sized hospital.

\section{Results}
\subsection{Single-domain learning}
Figure~\ref{fig:fig3} shows the single-domain learning results.
\oursarc{} shows comparable or higher prediction performances, on average, across the 12 tasks than other frameworks using domain knowledge (+0.8\%P AUROC on average against all frameworks on all three datasets, Fig~\ref{fig:fig3} (a) circle marks)
Appendix~\ref{apd:benchmark} provides the comparison results of \oursarc{} with a baseline involving more feature engineering.

\subsection{Pooled learning}
The results reveal that \oursarc{} exhibits a significant improvement in pooled learning when trained simultaneously on all three datasets, outperforming all other frameworks (+1.2\%P, Figure~\ref{fig:fig3} (a) triangles). This highlights the advantages of \oursarc{} which utilizes the textual representations of all features.
Compared with single-domain learning results, text-based embedding models (DescEmb and \oursarc{}) consistently demonstrate higher performances when trained on pooled datasets from all three sources. 
In contrast, conventional embedding models (SAnD and Rajkomar) show decreased or unchanged performances for pooled learning. 
In addition, for text-based embedding models, \oursarc{} outperforms DescEmb in most cases when all three data sources are pooled together.

\begin{table*}[ht]
\caption{\label{tab:pretrain} Self-supervised pre-training results}
\centering

\begin{tabular}{cc|c|ccc|ccc|} 
\cline{3-9}
\multicolumn{1}{l}{}                              & \multicolumn{1}{l|}{}                      & Structure & \multicolumn{3}{c|}{Hierarchical}                                                                                 & \multicolumn{3}{c|}{Flatten}                                                                           \\ \hline
\multicolumn{1}{|c|}{FT Source}                   & \multicolumn{1}{l|}{FT data size} & PT Source & \multicolumn{1}{c|}{Rand.Init} & \multicolumn{1}{c|}{SimCLR}             & Wav2vec                        & \multicolumn{1}{c|}{Rand.Init} & \multicolumn{1}{c|}{MLM}       & SpanMLM                         \\ \hline \bottomrule
\multicolumn{1}{|c|}{}                            &                                            & single    &                                & { 0.752***}       & { 0.733*}  &                                & 0.716                          & 0.709                           \\
\multicolumn{1}{|c|}{}                            & \multirow{-2}{*}{Sample data (10\%)}          & multi-source    & \multirow{-2}{*}{0.721}        & { {\bf 0.769***(0.006)}} & { 0.74**}  & \multirow{-2}{*}{0.711}        & 0.709                          & { 0.701*}   \\ \cline{2-9} 
\multicolumn{1}{|c|}{}                            &                                            & single    &                                & { 0.799**}        & 0.779                          &                                & { 0.773*}  & { 0.757**}  \\
\multicolumn{1}{|c|}{}                            & \multirow{-2}{*}{Sample data (30\%)}          & multi-source    & \multirow{-2}{*}{0.783}        & { {\bf 0.805***(0.008)}} & 0.781                          & \multirow{-2}{*}{0.767}        & 0.77                           & { 0.755**}  \\ \cline{2-9} 
\multicolumn{1}{|c|}{}                            &                                            & single    &                                & 0.83                                  & 0.832                          &                                & 0.805                          & { 0.799*}   \\
\multicolumn{1}{|c|}{\multirow{-6}{*}{MIMIC-III}} & \multirow{-2}{*}{Full Data}                & multi-source    & \multirow{-2}{*}{0.831}        & { {\bf 0.840**(0.011)}}  & 0.832                          & \multirow{-2}{*}{0.804}        & { 0.812**} & 0.801                           \\ \hline \bottomrule
\multicolumn{1}{|c|}{}                            &                                            & single    &                                & { 0.768***}       & { 0.737**} &                                & { 0.72*}   & { 0.695**}  \\
\multicolumn{1}{|c|}{}                            & \multirow{-2}{*}{Sample data (10\%)}          & multi-source    & \multirow{-2}{*}{0.721}        & { {\bf 0.771***(0.008)}} & { 0.734**} & \multirow{-2}{*}{0.708}        & { 0.725**} & { 0.689***} \\ \cline{2-9} 
\multicolumn{1}{|c|}{}                            &                                            & single    &                                & 0.793                                 & 0.785                          &                                & 0.77                           & { 0.757**}  \\
\multicolumn{1}{|c|}{}                            & \multirow{-2}{*}{Sample data (30\%)}          & multi-source    & \multirow{-2}{*}{0.789}        & { {\bf 0.801**(0.049)}}  & 0.79                           & \multirow{-2}{*}{0.773}        & { 0.78*}   & 0.768                           \\ \cline{2-9} 
\multicolumn{1}{|c|}{}                            &                                            & single    &                                & 0.818                                 & 0.818                          &                                & 0.801                          & 0.797                           \\
\multicolumn{1}{|c|}{\multirow{-6}{*}{eICU}}      & \multirow{-2}{*}{Full Data}                & multi-source    & \multirow{-2}{*}{0.817}        & {\bf 0.82(0.118)}                            & 0.818                          & \multirow{-2}{*}{0.802}        & 0.805                          & 0.796*                          \\ \hline \bottomrule
\multicolumn{1}{|c|}{}                            &                                            & single    &                                & { 0.729***}       & { 0.717**} &                                & { 0.711**} & { 0.677***} \\
\multicolumn{1}{|c|}{}                            & \multirow{-2}{*}{Sample data (10\%)}          & multi-source    & \multirow{-2}{*}{0.707}        & { {\bf 0.735***(0.008)}} & { 0.716**} & \multirow{-2}{*}{0.698}        & { 0.709**} & { 0.688*}   \\ \cline{2-9} 
\multicolumn{1}{|c|}{}                            &                                            & single    &                                & { 0.776**}        & 0.765                          &                                & 0.751                          & { 0.744*}   \\
\multicolumn{1}{|c|}{}                            & \multirow{-2}{*}{Sample data (30\%)}          & multi-source    & \multirow{-2}{*}{0.761}        & { {\bf 0.782***(0.009)}} & 0.764                          & \multirow{-2}{*}{0.753}        & { 0.757*}  & 0.751                           \\ \cline{2-9} 
\multicolumn{1}{|c|}{}                            &                                            & single    &                                & 0.837                                 & 0.834                          &                                & 0.815                          & { 0.808*}   \\
\multicolumn{1}{|c|}{\multirow{-6}{*}{MIMIC-IV}}  & \multirow{-2}{*}{Full Data}                & multi-source    & \multirow{-2}{*}{0.834}        & { {\bf 0.842**(0.014)}}  & 0.835                          & \multirow{-2}{*}{0.814}        & 0.817                          & { 0.809*}   \\ \hline
\end{tabular} %
\\

\raggedright \footnotesize{\hspace{1.3cm} p-value *:<0.1, **: <0.05, ***: <0.01}
\vskip -5pt 
\end{table*}

\subsection{Transfer learning}
Figure~\ref{fig:fig4} presents the transfer learning results.
To evaluate how the performance of the frameworks varies with the target dataset size, we use different proportions of the target dataset: x=0.0 indicates zero-shot learning, x=0.5 means fine-tuning with half of the target dataset, and x=1.0 is for fine-tuning with the entire target dataset.
For zero-shot learning, the text-based embedding methods (DescEmb and \oursarc) consistently outperform the code-based embedding methods (SAnD and Rajikomar) across all source and target pairs.

\oursarc{} demonstrate predominantly higher performance than the other models in most cases (+2.6\%P, red line over other lines). 
As the sample size of the target dataset decreases, the strength of \oursarc{} becomes more apparent (+12.5\%P, performance at x=0.0).
In further fine-tuning on the full dataset (marked with 1 on the x-axis), the code-based embedding models perform worse than \oursarc{} with single-domain learning performance (dotted line) in most cases. 
In contrast, \oursarc{} exhibits comparable or higher performance than single-domain learning, except when the model is trained on MIMIC-III and transferred to the eICU.
Next, we introduce two additional baselines capable of automatically map different code systems between two EHR datasets using unsupervised learning.
\oursarc{} exhibits a higher performance against unsupervised learning methods for code mapping, as shown in Table~\ref{tab:transfer}.

\subsection{Self-supervised pretraining}
Table ~\ref{tab:pretrain} presents the results.
Pretraining sources (PT Srouce) are in two settings, single(same as the fine-tune dataset) and multi-source (MIMIC-III+eICU+MIMIC-IV). Fine-tune(FT) data size are varied with the data sampled size (10\%, 30\%, full).
$\star$ indicates p-value from the t-test between the randomly initialized model and pretrained model results.
The highest performance for each fine-tune source, corresponding to the size of the fine-tune data, is highlighted in bold, and its p-value is indicated in parentheses.

The results show that \oursarc{} coupled with self-supervised pretraining methods (except SpanMLM) improves the prediction performance in most cases compared to models without pretraining.
Among the pretraining methods, SimCLR consistently outperforms the others, exhibiting the highest prediction AUROC, for both the sample-data and full-data scenarios. 
In particular, SimCLR exhibits an average increase of 0.1\%P and 0.6\%P in the AUROCs for the single- and multi-source pretraining, respectively.
The sample data results show that when the quantity of pretraining data exceeds that of the fine-tuning data to a larger extent, pretraining significantly affects the predictive downstream tasks.

\section{Discussion} \label{sec:discussion}
In this work, we addressed the dual challenges of multi-task prediction models for heterogeneous EHRs by proposing and investigating \oursarc{} for single-domain learning, pooled learning, transfer learning, and self-supervised pretraining. The results show that \oursarc{} achieves comparable or higher performances without relying on medical domain knowledge and by simply using all features as textual descriptions. 

In particular, for single domain learning, a comparison between \oursarc{} and Rajkomar suggests that assigning unique embeddings to all feature names and values is unnecessary, since treating them as textual descriptions leads to a comparable performance. 
Moreover, a comparison between \oursarc{} and DescEmb implies that \oursarc{} can better capture the underlying semantics of distinct EHR sources than DescEmb utilizing all available information in a medical event.
That is, applying medical domain knowledge to select a subset of meaningful features does not necessarily lead to a higher performance compared with simply using all possible features.
Overall, the single-domain learning results show that \oursarc{} achieves comparable or higher performances, even without relying on medical-domain knowledge, by simply using all features as textual descriptions. 
The improved AUROC achieved without significant feature engineering made this evident.

In the pooled learning, both text-based embedding models (DescEmb and GenHPF) significantly improved the prediction performance compared with conventional code-based embedding models.
This improvement results from the MIMIC and eICU datasets not sharing codes and from training conventional code-based embedding models on the pooled dataset expanding the number of required embeddings for each feature name and value, thereby preventing the models from leveraging larger amounts of training data.
Conversely, text-based embedding models can take advantage of the extensive volume of various sources since the sub-words of medical descriptions are common, even among entirely dissimilar EHR systems.
Furthermore, even within the text-based embedding models, \oursarc{} outperforms DescEmb in most cases, although DescEmb uses manually selected features from each dataset.
This highlights the advantage of \oursarc{} because it does not rely on any domain knowledge but rather uses all features in a textual form regardless of the EHR schema used.

In transfer learning, we observe a pattern similar to that in pooled learning; text-based embedding models consistently outperform code-based embedding methods.
Through these experiments, we demonstrate that GenHPF effectively resolves two challenges (multi-task learning, multi-source learning).
For multi-task learning, GenHPF outperforms models SAND and DescEmb, which employ feature selection by utilizing domain knowledge.
Regarding multi-source learning, GenHPF demonstrated better performance than conventional embedding models such as Rajkomar and SAnD. 

The self-supervised pretraining results show that SimCLR consistently outperforms the other methods.
We conjecture that SimCLR's pretraining process effectively facilitates prediction in downstream tasks by learning patient-level representations, whereas the other pretraining methods focus on learning either token-level or event-level representations within the same patient.
Furthermore, the performance improvement of \oursarc{} with multi-source pretraining provides insights into the necessity of pretraining on the pooled heterogeneous EHRs, which we believe is essential for large-scale EHR modeling.

Implementing GenHPF in a real-world hospital requires appropriate hardware resources, including GPUs connected to EHR database. 
Once operational, the framework minimally preprocesses patient data for various prediction tasks. 
A key advantage of GenHPF is that it can be integrated into any EHR system without requiring specific modifications, thereby significantly reducing both time and implementation costs. 
However, this approach to minimal preprocessing results in a larger input size, requiring higher computational requirements.

\section{limitation}
Although \oursarc{} demonstrated promising results, it still has limitations. 
First, since \oursarc{} utilizes as many features as possible from EHR events, computational constraints must be considered. 
Therefore, we used a subset of EHR events (lab tests, prescriptions, and input events) in this work.
Better performance is expected if we exploit all EHR event types using more memory-efficient models~\cite{choromanski2020rethinking, gu2022efficiently}.

Second, as the current framework for multi-source learning relies on textual representation, it is limited to EHRs that share the same language. 
Lastly, we used only tabular data in the EHR; thus, future studies should consider incorporating additional modalities (e.g., radiographic images) into the framework.

\section{conclusion}
In conclusion, our study illustrates the potential of \oursarc{} for various learning scenarios, including single-domain, pooled, transfer learning, and self-supervised pretraining. 
The effectiveness of the framework without relying on medical domain knowledge and its ability to capture the underlying semantics of distinct EHR sources make it a promising approach for large-scale EHR modeling in the future.
Furthermore, With the advent of large language models (LLMs) such as Chat-GPT, feeding text-based EHRs into an LLM via the GenHPF framework (with its ability to handle any EHR in text form) would allow for EHR predictions, either by fine-tuning the LLM or using the in-context learning technique.
This would open up a wide set of applications that could reduce complications and improve patient care with less reliance on EHR schemas and feature engineering, such as predicting patient outcomes, intervention, and personalizing patient care.

\section*{Acknowledgment}
This work was (partially) supported by the KAIST-NAVER Hyper-Creative AI Center, Institute of Information \& Communications Technology Planning \& Evaluation (IITP) grant (No.2019-0-00075), National Research Foundation of Korea (NRF) grant (NRF-2020H1D3A2A03100945), Korea Medical Device Development Fund grant (Project Number: 1711138160, KMDF\_PR\_20200901\_0097), and Korea Health Industry Development Institute (KHIDI) grant (No.HR21C0198), funded by the Korea government (MSIT, MOTIE, MOHW, MFDS).

\appendices
\section{Supplementary Results}
\subsection{Comparison of \oursarc{} with Benchmark~\cite{mcdermott2021a}} \label{apd:benchmark}
We compare \oursarc{} with Benchmark~\cite{mcdermott2021a}, shown in table~\ref{tab:benchmark}.
The best performances for each dataset are in bold.
While Benchmark offers an expert-designed, feature-engineered prediction pipeline, comparing it with \oursarc{} allows us to assess the effectiveness of our method, which operates without domain-specific knowledge.
Benchmark originally used all tables, including lab tests and chart events. Due to the high computational demands from numerous chart events, we limited our comparison to the lab test table. This ensures a fair comparison, as both our method and Benchmark share only the lab test event.
\oursarc{} generally exhibits a higher performance than that of Benchmark in most prediction tasks.
\vskip -10pt
\begin{table}[ht]
\centering
\scriptsize
\caption{\label{tab:benchmark} Comparison with benchmark model (only lab features)}
\begin{tabular}{c|ccc}
\hline
Source    & Benchmark & \multicolumn{1}{l}{Rajikomar} & GenHPF (ours) \\ \hline 
MIMIC-III & 0.779       & \textbf{0.786**(0.031)}                & 0.784*          \\
eICU      & 0.783       & 0.788*                          & \textbf{0.79**(0.024)} \\
\hline
\end{tabular} %
\vskip -5pt 
\end{table}

\subsection{Comparison \oursarc{} with unsupervised learning methods in transfer learning}
AutoMap~\cite{wu2022automap} and Muse~\cite{conneau2017word} use the same model architecture as Rajikomar~\cite{rajkomar2018scalable} but can leverage learned embedding through the unsupervised pretraining of code features between the source and target datasets.
We use these baselines for fair a comparison when transferring code-based embedding models, giving pretrained embedding, not just randomly initialized.
Results are shown in Table ~\ref{tab:transfer}.
The two unsupervised learning methods for code-mapping do not exhibit improvement over Rajkomar in the full dataset performance.
This indicates that pretraining with code-mapping between two sources using different EHR code schemes does not yield a performance improvement, and the original paper did not conduct the experiments across different EHRs.
However, \oursarc{} which utilizes text-based embedding outperforms the baselines (AutoMap, Muse, and Rajikomar) in both zero-shot learning and full dataset fine-tuning.
\vskip -10pt

\begin{table}[h]
\centering
\scriptsize
\caption{\label{tab:transfer} Transfer learning with additional baselines}
\begin{tabular}{c|cllll}
\hline
Source -\textgreater Target                       & \multicolumn{1}{l}{} & AutoMap  & MUSE     & Rajikomar & GenHPF \\ \hline
\multirow{2}{*}{MIMIC-IV -\textgreater eICU}      & Zero-shot            & 0.473*** & 0.502*** & 0.505***  & 0.610  \\
                                                  & Finetune             & 0.811*   & 0.812*   & 0.815     & 0.821  \\ \hline
\multirow{2}{*}{eICU -\textgreater MIMIC-IV}      & Zero-shot            & 0.531*** & 0.52***  & 0.535***  & 0.611  \\
                                                  & Finetune             & 0.829**  & 0.831*   & 0.836     & 0.839  \\ \hline
\multirow{2}{*}{MIMIC-IV -\textgreater MIMIC-III} & Zero-shot            & 0.509*** & 0.524*** & 0.525***  & 0.729  \\
                                                  & Finetune             & 0.817*** & 0.826**  & 0.832*    & 0.838  \\ \hline
\end{tabular}
\end{table}

\section{Statistics for prediction tasks}\label{apd:data}
This section presents the statistics for the prediction tasks.
All numbers represent the composition ratios as percentages.
\noindent{\textbf{Binary Classification Tasks Table~\ref{tab:binary}}} The tasks include predicting mortality, long-term mortality, los3, los7, and readmission. 

\noindent{\textbf{Multi-class Classification Tasks Table~\ref{tab:multi-class}}} The tasks include predicting the final acuity, imminent discharge, and several lab values (creatinine, bilirubin, platelets, and WBC). 
For laboratory values, 'Null' denotes ICU samples that involve dialysis.
The loss is not computed for this null class during training phases. 
For the final acuity and imminent discharge, samples outside the predefined classes are marked as 'Null'. 

\noindent{\textbf{Multi-label Classification Tasks Table~\ref{tab:multi-label}}} Each row represents a class label, and the corresponding percentages denote the proportion of instances assigned to each class in the respective dataset.
Class unification across the datasets follows DescEmb~\cite{hur2022unifying}.

\begin{table}[ht]
    \caption{\label{tab:binary} Statistics for binary classification tasks}
    \centering
    \scriptsize
    \begin{tabular}{lrrrrrr}
    \hline
    
                          & \multicolumn{2}{c}{MIMIC-III} & \multicolumn{2}{c}{eICU} & \multicolumn{2}{c}{MIMIC-IV} \\ 
                          \hline
    Task/Class            & 0             & 1             & 0           & 1          & 0             & 1            \\ \hline
    mortality             & 98.3          & 1.7           & 98.4        & 1.6        & 98.4          & 1.6          \\
    long\_term\_mortality & 91.6          & 8.4           & 92.8        & 7.2        & 92.3          & 7.7          \\
    los\_3day             & 65.8          & 34.2          & 71.2        & 28.8       & 70.8          & 29.2         \\
    los\_7day             & 87.9          & 12.1          & 91.2        & 8.8        & 90.9          & 9.1          \\
    readmission           & 94.5          & 5.5           & 90          & 10         & 92.7          & 7.3          \\ \hline
    \end{tabular}

\end{table}
\begin{table}[ht]
    \caption{\label{tab:multi-class} Statistics for multi-class classification tasks
    }
    \centering
    \scriptsize
\begin{tabular}{c|c|ccc}
\hline
Task                                 & Class & MIMIC-III & eICU & MIMIC-IV \\ \hline
\multirow{7}{*}{final\_acuity}       & Null  & 0.0       & 1.0  & 0.8      \\
                                     & 0     & 51.9      & 58.5 & 51.5     \\
                                     & 1     & 3.4       & 3.4  & 3.7      \\
                                     & 2     & 7.2       & 5.0  & 6.0      \\
                                     & 3     & 9.6       & 13.2 & 11.5     \\
                                     & 4     & 12.4      & 4.6  & 7.3      \\
                                     & 5     & 15.5      & 13.6 & 18.6     \\ \hline
\multirow{7}{*}{imminent\_discharge} & Null  & 0.0       & 0.1  & 0.5      \\
                                     & 0     & 95.4      & 1.6  & 1.6      \\
                                     & 1     & 0.2       & 5.5  & 2.9      \\
                                     & 2     & 1.7       & 90.9 & 93.9     \\
                                     & 3     & 0.5       & 1.6  & 0.5      \\
                                     & 4     & 0.1       & 0.0  & 0.0      \\
                                     & 5     & 0.1       & 0.1  & 0.1      \\ \hline
\multirow{6}{*}{creatinine}          & Null  & 15.7      & 25.1 & 14.8     \\
                                     & 0     & 59.2      & 49.8 & 58.8     \\
                                     & 1     & 16.1      & 15.1 & 16.4     \\
                                     & 2     & 6.2       & 6.5  & 6.4      \\
                                     & 3     & 1.6       & 1.9  & 1.9      \\
                                     & 4     & 0.1       & 1.6  & 1.8      \\ \hline
\multirow{6}{*}{bilirubin}           & Null  & 78.5      & 71.4 & 74.0     \\
                                     & 0     & 13.2      & 21.9 & 16.4     \\
                                     & 1     & 2.8       & 3.3  & 3.4      \\
                                     & 2     & 3.6       & 2.5  & 4.1      \\
                                     & 3     & 1.0       & 0.5  & 1.1      \\
                                     & 4     & 0.8       & 0.3  & 0.9      \\ \hline
\multirow{6}{*}{platelets}           & Null  & 11.9      & 25.2 & 11.7     \\
                                     & 0     & 61.8      & 49.4 & 58.0     \\
                                     & 1     & 16.9      & 15.9 & 19.1     \\
                                     & 2     & 7.5       & 7.2  & 8.6      \\
                                     & 3     & 1.7       & 1.6  & 2.2      \\
                                     & 4     & 0.3       & 0.3  & 0.4      \\ \hline
\multirow{4}{*}{wbc}                 & Null  & 12.3      & 24.7 & 11.8     \\
                                     & 0     & 3.6       & 45.5 & 3.9      \\
                                     & 1     & 53.5      & 26.6 & 55.0     \\
                                     & 2     & 31.0      & 2.6  & 29.4     \\ \hline
\end{tabular}
\end{table}
\begin{table}[ht]
    \caption{\label{tab:multi-label} Statistics for multi-label classification task(Dx)}
    \centering
    \scriptsize
    \begin{tabular}{rrrr}
    \hline
    \multicolumn{1}{l}{class} & \multicolumn{1}{l}{MIMIC-III} & \multicolumn{1}{l}{eICU} & \multicolumn{1}{l}{MIMIC-IV} \\ \hline
    \multicolumn{1}{r|}{0}    & 4.79\                        & 5.35\                   & 4.67\                       \\
    \multicolumn{1}{r|}{1}    & 4.29\                        & 2.27\                   & 4.06\                       \\
    \multicolumn{1}{r|}{2}    & 11.53\                       & 10.98\                  & 10.40\                      \\
    \multicolumn{1}{r|}{3}    & 6.33\                        & 4.65\                   & 6.76\                       \\
    \multicolumn{1}{r|}{4}    & 6.02\                        & 4.48\                   & 7.83\                       \\
    \multicolumn{1}{r|}{5}    & 5.10\                        & 6.10\                   & 6.14\                       \\
    \multicolumn{1}{r|}{6}    & 13.62\                       & 22.09\                  & 11.26\                      \\
    \multicolumn{1}{r|}{7}    & 8.15\                        & 13.73\                  & 6.88\                       \\
    \multicolumn{1}{r|}{8}    & 7.45\                        & 5.87\                   & 7.23\                       \\
    \multicolumn{1}{r|}{9}    & 7.37\                        & 8.67\                   & 6.93\                       \\
    \multicolumn{1}{r|}{10}   & 0.06\                        & 0.16\                   & 0.08\                       \\
    \multicolumn{1}{r|}{11}   & 1.75\                        & 0.65\                   & 1.51\                       \\
    \multicolumn{1}{r|}{12}   & 3.73\                        & 0.68\                   & 4.36\                       \\
    \multicolumn{1}{r|}{13}   & 0.56\                        & 0.02\                   & 0.59\                       \\
    \multicolumn{1}{r|}{14}   & 7.29\                        & 6.16\                   & 5.72\                       \\
    \multicolumn{1}{r|}{15}   & 4.68\                        & 5.08\                   & 6.64\                       \\
    \multicolumn{1}{r|}{16}   & 7.30\                        & 3.05\                   & 8.95\                       \\ \hline
    \end{tabular}

\end{table}

\end{document}